\documentclass{article}

\PassOptionsToPackage{numbers,compress}{natbib}

\usepackage[main, final]{neurips_2026}

\usepackage[utf8]{inputenc} 
\usepackage[T1]{fontenc}    
\usepackage{hyperref}       
\usepackage{url}            
\usepackage{booktabs}       
\usepackage{amsfonts}       
\usepackage{nicefrac}       
\usepackage{microtype}      
\usepackage{xcolor}         

\usepackage{booktabs}
\usepackage{makecell}
\hypersetup{hidelinks}

\title{A Control-Theoretic Approximation to \\ Predictive Coding Dynamics}

\author{%
Ryan Fayyazi\footnotemark[1] \\
New York University \\
\texttt{ryan.fayyazi@nyu.edu} \\
\And
Kyle Daruwalla\thanks{Equal contribution.} \\
Cold Spring Harbor Laboratory \\
\texttt{daruwal@cshl.edu} \\
\And
Mitra Javadzadeh \\
Cold Spring Harbor Laboratory \\
\texttt{javadzadeh@cshl.edu} \\
}

\usepackage{graphicx}
\usepackage{amsmath}
\usepackage{amssymb}
\usepackage{algorithm}
\usepackage{wrapfig}
\usepackage{cleveref}
\usepackage[separate-uncertainty=true]{siunitx}
\crefname{section}{section}{sections}
\Crefname{section}{Section}{Sections}
\crefname{figure}{figure}{figures}
\Crefname{figure}{Figure}{Figures}
\newcommand{\vv}[1]{\mathbf{#1}}
\newcommand{\mm}[1]{\mathrm{\mathbf{#1}}}
\newcommand{\deriv}[2]{\frac{\mathrm{d} #1}{\mathrm{d} #2}}

\begin{document}

\maketitle
\setcounter{footnote}{0}

\begin{abstract}
Backpropagation remains the dominant method for training deep neural networks, but its reliance on non-local computations has motivated the search for biologically plausible alternatives. Two prominent frameworks, predictive coding (PC) and deep feedback control (DFC), offer distinct approaches to local learning, yet their relationship remains unclear. Here, we show that DFC arises as a local approximation to predictive coding dynamics near fixed points, with exact equivalence in shallow linear networks and systematic deviations arising with depth, nonlinearity, and distance from equilibrium. 
We characterize how this approximation, which we call Approx PC, deviates from PC in its activity dynamics, and show that these deviations can be mitigated by relinearizing the feedback controller and approaching the output target through a sequence of intermediate nudges. In networks trained on MNIST and FashionMNIST, Approx PC reaches task performance comparable to PC trained with the same nudges, and weights learned with it transfer directly to PC inference.
Our results unify energy-based and control-theoretic perspectives on learning and open the door to applying tools from control theory to the training of predictive coding networks.
\end{abstract}

\section{Introduction}

Backpropagation has become the dominant algorithm for training deep neural networks, yet it relies on computational primitives that are difficult to reconcile with biological neural circuits, including weight transport across layers \citep{akrout2019}, global error signals \citep{oreilly1996, sacramento2018}, and phased learning stages \citep{jaderberg2017}. The brain, on the other hand, relies on synaptic plasticity, which is widely believed to depend on local information available at each synapse, such as pre- and post-synaptic activity and modulatory feedback, and does not require distinct forward and backward passes \citep{OrorbiaBrainInspiredMachineIntelligence2023}. This contrast has motivated a large body of work seeking biologically plausible alternatives for credit assignment in deep neural networks \citep{BalduzziKickbackCutsBackprop2015,CraftonDirectFeedbackAlignment2019,LillicrapRandomFeedbackWeights2014,ScellierEquilibriumPropagationBridging2017,PogodinKernelizedInformationBottleneck2020,DaruwallaInformationBottleneckbasedHebbian2024,FremauxNeuromodulatedSpikeTimingDependentPlasticity2016,LillicrapBackpropagationBrain2020,lee2015,meulemans2020}. Among these, two influential approaches have emerged from distinct perspectives: (1) predictive coding (PC) networks \citep{rao1999, friston2005, whittington2017,millidge2022}, inspired by sensory processing in cortical circuits and typically formulated as energy-based models that minimize a global objective by reducing local mismatch between bottom-up and top-down signals, and (2) deep feedback control (DFC) \citep{MeulemansCreditAssignmentNeural2021}, motivated by principles of motor control and rooted in control-theoretic frameworks, where credit assignment is achieved via dynamical error correction through a feedback controller.

\begin{figure}[!ht]
\centering
\includegraphics[width=0.9\linewidth]{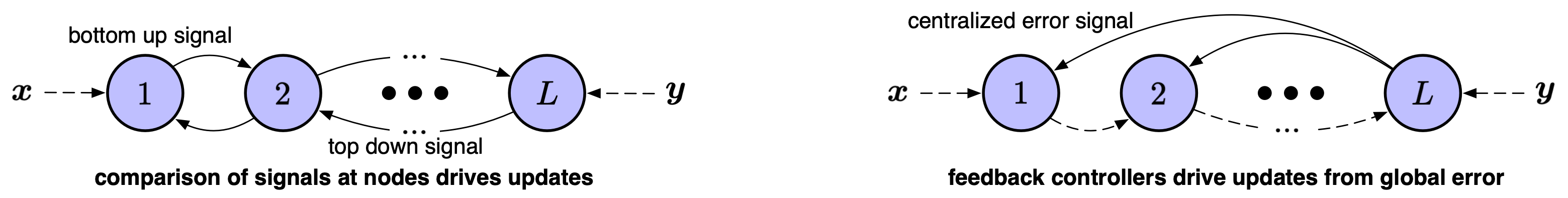}
\caption{Illustrations of error signal sources in predictive coding (left) and deep feedback control (right). PC errors are distributed while DFC is centralized. Pathways involved in calculating errors are denoted by solid lines, while general information pathways are denoted by dashed lines.}
\label{fig:pc-vs-dfc}
\end{figure}

While both frameworks implement local learning rules, their underlying neural dynamics appear markedly different. Although both rely on error signals, their origin and role differ: in PC, errors are computed locally at each layer, whereas in DFC they arise from a global target and are distributed through the network via the controller (see \Cref{fig:pc-vs-dfc}). This apparent difference raises the question of whether these two classes of models implement fundamentally distinct mechanisms for credit assignment, or whether they can be understood within a common framework.

From a biological perspective, the connectivity motifs required by both frameworks are present in the cortex: hierarchical organization with local feedforward–feedback loops supports predictive coding, while long-range feedback from higher-order to early sensory areas enables global control signals. This suggests that the brain may support both local error correction mechanisms and more global control-driven dynamics, raising the possibility that the two frameworks reflect different regimes of a shared underlying computation.

Here, we show that DFC can be interpreted as a local approximation to PC dynamics in the vicinity of fixed points. This equivalence is exact for shallow linear networks and progressively breaks down with increasing network depth, with the introduction of nonlinearities, and as the system moves further from equilibrium. Building on this connection, we propose a practical scheme to approximate PC using DFC by decomposing the trajectory toward a target into a sequence of intermediate subgoals. Within each subgoal, DFC dynamics drive the network toward a local equilibrium before transitioning to the next stage. This perspective suggests that distributed, layer-wise error signals can be effectively replaced by a centralized feedback controller that sequentially guides the network through intermediate targets. \textbf{Our contributions are organized in three parts:}\\ \\
\emph{Theory:} We establish a formal connection between predictive coding and deep feedback control by showing that PC activity dynamics can be written as a distributed feedback controller, which reduces near fixed points to a DFC-style controller driven by the output error. We call the resulting approximate dynamics Approx PC. In linear networks, we show that the two are exactly equivalent with a single hidden layer, and that with two hidden layers they share the same fixed point and differ only by an additional damping term in PC. \\ \\
\emph{Characterizing the approximation:} Through controlled probes of activity dynamics on single inputs, we show how Approx PC deviates from PC as perturbation size, depth, and nonlinearity increase. We demonstrate that these deviations are mitigated by relinearizing the feedback controller or by approaching the output target through a sequence of intermediate nudges. \\ \\
\emph{Validation under learning:} We test whether the approximation error compounds during training and find that it does not substantially affect what the network learns. On MNIST \citep{lecun1998} and FashionMNIST \citep{XiaoFashionMNISTNovelImage2017}, networks trained with Approx PC reach task performance comparable to that of PC trained with the same nudges, their learned weights transfer directly to PC inference, and their hyperparameter sensitivity is similar to that of PC.\footnote{Code is available at \url{https://github.com/darsnack/pc-dfc}.}

\section{Related work}

Although not the focus of this work, many alternative biologically plausible learning rules use local signals and objectives to learn \citep{ScellierEquilibriumPropagationBridging2017,PogodinKernelizedInformationBottleneck2020,DaruwallaInformationBottleneckbasedHebbian2024} or various centralized feedback pathways to drive learning \citep{BalduzziKickbackCutsBackprop2015,CraftonDirectFeedbackAlignment2019,LillicrapRandomFeedbackWeights2014,PogodinKernelizedInformationBottleneck2020,DaruwallaInformationBottleneckbasedHebbian2024,FremauxNeuromodulatedSpikeTimingDependentPlasticity2016}. Many of these have theoretical relationships to both PC and DFC, but our objective is to understand PC through a specific family of control-based rules. In particular, while recent hierarchical variants of PC incorporate long-range feedback pathways, here we focus on the standard PC formulation. The work most closely related to ours is \citep{MeulemansLeastcontrolPrincipleLocal2022}, where the authors briefly note that PC can be understood as a solution to a least-control problem. Yet the controller they propose is distributed rather than a DFC-style feedback controller driven by the final layer's error. We discuss these distinctions in more detail in \Cref{theory}.

\section{Background}

Consider a network comprising $L$ layers of neurons, with input $\vv{x}\left(t\right) \in \mathbb{R}^{d}$. Let the membrane potential of neurons in layer $\ell$ be $\vv{v}^{\ell}\left(t\right)$, and let their activity be $\vv{r}^{\ell}\left(t\right) = \phi(\vv{v}^{\ell}(t))$, where $\phi$ is an activation function. Define $\vv{v}^{0}\left(t\right):= \vv{x}\left(t\right)$ as the input. The parameters connecting layer $\ell-1$ and $\ell$ of the network are denoted $\mm{W}^{\ell}\left(t\right)$. In general, we will seek to optimize both the potentials $\vv{v}^{\ell}$ and the parameters $\mm{W}^{\ell}$ concurrently. For concision, variables with $t$ in the subscript denote time-varying functions, i.e., $\vv{x}_{t}:= \vv{x}\left(t\right)$.

\subsection{Predictive coding}
\label{pc-background}

Given a network as defined above, predictive coding seeks to minimize the network's total energy, defined as the sum of layer-wise energies. The energy of a given layer is defined as the difference between its potential and the input it receives from the previous layer \footnote{The term in Eq. \eqref{eq:pc-energy} associated with the final layer’s energy corresponds approximately to an MSE loss, but there is a more general formulation which includes a generic loss by adding a term $\mathcal{L}\left(\vv{v}_{t}^{L}, \vv{y}_{t}\right)$ to the total energy \citep{pinchetti2022}.}:
\begin{equation}
E_{t} = \frac{1}{2} \sum_{\ell} \|\underbrace{\vv{v}_{t}^{\ell} - \mm{W}_{t}^{\ell} \vv{r}_{t}^{\ell - 1}}_{\vv{e}_{t}^{\ell}}\|_{2}^{2}
\label{eq:pc-energy}
\end{equation}
To minimize the energy, the potential and weight dynamics at every layer follow the negative gradient,
\begin{equation}
\begin{aligned}
\tau_{v} \deriv{\vv{v}_{t}^{\ell}}{t} &= -\nabla_{\vv{v}_{t}^{\ell}} E_{t} &= \begin{cases} - \vv{e}_{t}^{\ell} + \phi'\left(\vv{v}_{t}^{\ell}\right) \odot \left(\mm{W}_t^{\ell+1}\right)^{\top} \vv{e}_{t}^{\ell+1} &\text{if } \ell<L \\ - \vv{e}_{t}^{L} &\text{if } \ell = L \end{cases}
\end{aligned}
\label{eq:pc-potential-dynamics}
\end{equation}
\begin{equation}
\tau_{w} \deriv{\mm{W}_{t}^{\ell}}{t} = -\nabla_{\mm{W}_{t}^{\ell}} E_{t} = \vv{e}_{t}^{\ell} \left(\vv{r}_{t}^{\ell-1}\right)^{\top}
\label{eq:pc-weight-dynamics}
\end{equation}
where $\tau_{v}$ is the neural time constant and $\tau_{w}$ is the weight time constant. Thus, the network activity at each layer attempts to align top-down and bottom-up signals, yielding representations that minimize the system's overall energy.

The network operates in two modes: \emph{inference} and \emph{learning}. In both modes, the input layer (i.e., $\vv{v}_{t}^{0}$) is \emph{clamped}, meaning it is fixed to an input $\vv{x}_{t}$. This, along with \eqref{eq:pc-potential-dynamics}, drives changes in the network activity during inference, while the weights are fixed. In supervised learning, we additionally clamp the final layer's activity to the labels $\vv{y}_{t}$. This, along with \eqref{eq:pc-potential-dynamics} and \eqref{eq:pc-weight-dynamics}, drives both network activity and parameter updates during learning.

Note that we adopt a forward prediction convention, in which each layer predicts the activity of the next layer. While much of the PC literature uses the opposite (top-down) convention~\citep{rao1999, whittington2017}, we choose the forward convention because it is more familiar to a general deep learning audience and matches the convention used for DFC~\citep{MeulemansCreditAssignmentNeural2021}.

\subsection{Deep feedback control}
\label{dfc-background}

Deep feedback control (DFC) \citep{MeulemansCreditAssignmentNeural2021,MeulemansMinimizingControlCredit2022,MeulemansLeastcontrolPrincipleLocal2022} is an alternative framework, motivated by control theory, for adjusting the network activity and parameter dynamics to produce optimal predictions. DFC consists of a dynamical multi-layer network, where each layer integrates feedforward inputs with a feedback control signal \footnote{Our notation differs slightly from \citep{MeulemansCreditAssignmentNeural2021} for consistency with the rest of the text.}:
\begin{equation}
\tau_{v} \deriv{\vv{v}_{t}^{\ell}}{t} = -\vv{v}_{t}^{\ell} + \mm{W}_{t}^{\ell} \vv{r}_{t}^{\ell - 1} + \vv{u}_{t}^{\ell}
\label{eq:dfc-dynamics}
\end{equation}
Following control theory, we define a target, $\vv{r}^{*}$, for the final layer activity. The feedback controller determines the control signal at each layer, $\vv{u}_{t}^{\ell}$, to minimize the error between final layer activity and the target, $\vv{e}_{t}^{L} = \vv{r}^{*} - \vv{r}_{t}^{L}$. For proportional control, which we consider for this study, the overall control signal is $\vv{u}_{t} = K_{p} \vv{e}_{t}^{L}$, which drives each layer through feedback weights, i.e., $\vv{u}_{t}^{\ell} = \mm{Q}^{\ell} \vv{u}_{t}$.

We choose $\vv{r}^{*}$ to minimize a loss function, $\mathcal{L}$,
\begin{equation}
\vv{r}^{*} := \vv{r}_{\text{ss}}^{L} - \left.\lambda \deriv{\mathcal{L}}{\vv{r}^{L}}\right|_{\vv{r}^{L} = \vv{r}_{\text{ss}}^{L}} = \vv{r}_{\text{ss}}^{L} + \delta^{L}
\label{eq:dfc-nudge}
\end{equation}
where $\vv{r}_{\text{ss}}^{L}$ is the steady state of $\vv{r}_{t}^{L}$ in the absence of control feedback (i.e., $\vv{u}_{t}^{L} = 0$). The \emph{nudge size}, $\lambda > 0$, is chosen to be small to ensure that $\vv{r}^{*}$ remains close to the steady-state activity.

While \eqref{eq:dfc-dynamics} ensures that the activity of each layer is modulated to bring the final layer close to the target, this might require a large control signal---$\|\vv{u}_t\| \gg 0$. DFC follows the principle of least control and seeks to adjust $\mm{W}^{\ell}$ to reduce $\|\vv{u}_{t}\|$. This is achieved through the local learning dynamics,
\begin{equation}
\tau_{w} \deriv{\mm{W}_{t}^{\ell}}{t} = \left(\vv{r}_{t}^{\ell} - \phi(\mm{W}_t^\ell \vv{r}_t^{\ell - 1})\right) \left(\vv{r}_{t}^{\ell - 1}\right)^{\top}
\label{eq:dfc-weight-dynamics}
\end{equation}
which seeks to minimize the mean squared error between a layer's activity $\vv{r}_{t}^{\ell}$ and the input from the previous layer, $\phi(\mm{W}_t^\ell \vv{r}_t^{\ell - 1})$. Note that when $\tau_w \gg \tau_v$, both \eqref{eq:dfc-dynamics} and \eqref{eq:dfc-weight-dynamics} can be applied synchronously. Additionally, one could adjust the feedback weights, $\mm{Q}^{\ell}$, to reduce $\|\vv{u}_{t}\|$, but in general a wide range of fixed feedback weights is sufficient.

\section{Theoretical analysis of predictive coding as deep feedback control}
\label{theory}

Taken at face value, the activity dynamics of predictive coding \eqref{eq:pc-potential-dynamics} and deep feedback control \eqref{eq:dfc-dynamics} appear distinct. Our aim is to show that these two frameworks are closely related---specifically, that PC can be approximated by a specific DFC controller. For the remainder of this section, we focus on the activity dynamics, which occur much faster than the weight dynamics ($\tau_v \ll \tau_w$). In general, both PC and DFC follow equivalent weight dynamics (see \eqref{eq:pc-weight-dynamics} and \eqref{eq:dfc-weight-dynamics}, which differ only in activity vs. potential space). We use PC weight dynamics in our final proposed scheme, with the local error $\vv{e}^\ell$ replaced by the control signal $\vv{u}^\ell$.

In PC, following \eqref{eq:pc-potential-dynamics}, the potentials of layers $\ell < L$ evolve according to (see \Cref{pc_activity_deriv} for more):
\begin{equation}
\tau_{v} \deriv{\vv{v}_{t}^{\ell}}{t} = -\vv{e}_{t}^{\ell} + \left(\mm{J}_{t}^{\ell+1}\right)^{\top} \vv{e}_{t}^{\ell+1}=-\vv{v}_{t}^{\ell} + \mm{W}_{t}^{\ell} \vv{r}_{t}^{\ell - 1} + \left(\mm{J}_{t}^{\ell + 1}\right)^{\top} \vv{e}_{t}^{\ell + 1}
\label{eq:derive-pc-dfc-dynamics-equiv}
\end{equation}
Thus, we see that PC activity dynamics are very similar to DFC dynamics in \eqref{eq:dfc-dynamics}, with the layer-wise control given by $\vv{u}_{t}^{\ell} = \left(\mm{J}_{t}^{\ell + 1}\right)^{\top} \vv{e}_{t}^{\ell + 1}$. Unlike DFC, this controller does not directly feed back the error signal from the output. Instead, it defines a recurrence where each layer's error controls the activity of the preceding layer. So, while the dynamics as expressed in \eqref{eq:derive-pc-dfc-dynamics-equiv} are similar, PC dynamics follow an entirely different control strategy.

For the last layer, PC drives activity by clamping the output, while in DFC, the controller computes an error signal, $\vv{r}^* - \vv{r}_t^L$. We reconcile these here by clamping the output to the target $\vv{r}^*$. Then for PC, we have that $\vv{e}_{t}^{L} = \vv{v}_{t}^{L} - \mm{W}_{t}^{L} \vv{r}_{t}^{L - 1} = \phi^{-1}\left(\vv{r}^{*}\right) - \mm{W}_{t}^{L} \vv{r}_{t}^{L - 1}$ and $\nabla_{\vv{v}_{t}^{L}} E_{t} = 0$ (since $\phi^{-1}\left(\vv{r}^{*}\right)$ is constant) \footnote{We assume for simplicity that $\phi$ is the identity at the final layer.}. Thus, the final layer control signal is $\vv{u}_{t}^{L} = 0$. During inference, $\vv{r}_{t}^{L}$ is not clamped and a non-zero error, $\vv{e}_{t}^{L} = \vv{u}_{t}^{L}$, drives the output dynamics.

This completes the control-theoretic perspective on PC, which captures its dynamics as a distributed, time-varying proportional feedback controller (see \Cref{fig:pc-controller}). Indeed, \citep{MeulemansLeastcontrolPrincipleLocal2022} show that under a specific objective, PC is the optimal solution to a least-control problem. We, however, are concerned not with any control strategy but specifically with those that match DFC, in which the error computed at the output modulates hidden layers directly through feedback control. Next, we will make approximations to the distributed controller near fixed points to make PC and DFC completely equivalent.

\subsection{Equivalent control strategies for PC and DFC at fixed points}

Once converged, the PC network is at a fixed point, so $\deriv{v_{t}^{\ell}}{t}$ = 0. In this regime, \eqref{eq:pc-potential-dynamics} can be rewritten as $\vv{e}_{t}^{\ell} = \left(\mm{J}_{t}^{\ell + 1}\right)^{\top} \vv{e}_{t}^{\ell + 1}$. This implies that for PC networks at a fixed point,
\begin{equation}
\vv{u}_{t}^{\ell} = \left(\mm{J}_{t}^{\ell + 1}\right)^{\top} \vv{e}_{t}^{\ell + 1} = \left(\prod_{k = \ell + 1}^{L} \mm{J}_{t}^{k}\right)^{\top} \vv{e}_{t}^{L},
\label{eq:pcn_fixedpoint-control-signal}
\end{equation}
so the $\vv{u}_{t}^{\ell} $ inputs to each layer depend only on the error in the last layer. The control signal in \eqref{eq:pcn_fixedpoint-control-signal} matches that of DFC, where the error is calculated at the final layer, and the feedback weights for each layer are given by:
\begin{equation}
\mm{Q}_{t}^{\ell} K_{p} := \left(\prod_{k = \ell + 1}^{L} \mm{J}_{t}^{k}\right)^{\top},
\label{eq:approx-dfc-control-signal}
\end{equation}
where $K_p \geq 0$ is the proportional control constant. Thus, PC at fixed points can be exactly expressed using the feedback control strategy employed by DFC (see \Cref{pc_activity_deriv}). However, the feedback weights are time-dependent. So, \emph{we make two approximations: first, that \eqref{eq:pcn_fixedpoint-control-signal} and \eqref{eq:approx-dfc-control-signal} hold locally near fixed points; second, that the time-dependent Jacobian can be approximated by linearizing the network dynamics around the fixed point.} The latter is only required for nonlinear networks, as the Jacobians for linear networks are constant with respect to time. \textbf{Our final approach, which we refer to as ``Approx PC'', is summarized by}
\begin{equation}
\tau_{v} \deriv{\vv{v}_{t}^{\ell}}{t} = -\vv{v}_{t}^{\ell} + \mm{W}_{t}^{\ell} \vv{r}_{t}^{\ell - 1} + \vv{u}_{t}^{\ell} \quad \text{where} \quad \vv{u}_{t}^{\ell} = \left(\prod_{k = \ell + 1}^{L} \mm{J}_{0}^{k}\right)^{\top} \vv{e}_{t}^{L}
\label{eq:approx-dfc-dynamics}
\end{equation}
where we assume the network is initially at a fixed point at $t = 0$.

\subsection{Approximation characterization in linear networks}
\label{theory_3layer}

Assume that the network is at steady state such that \eqref{eq:approx-dfc-control-signal} holds exactly. We then clamp the target to $\vv{r}^{*}$, which is selected to be close to the current steady-state output. This perturbs the system away from the fixed point by a small amount. There are multiple sources of error: \eqref{eq:pcn_fixedpoint-control-signal} may not hold, and for nonlinear networks, the Jacobians are only exact at the fixed point. To disentangle these sources of error, we begin by considering linear networks.

In a two-layer linear network (one hidden layer), we have activity dynamics for only one layer (the hidden layer), as the input and output layers are clamped. Since $\mm{Q}^{L-1} = (\mm{W}^L)^\top$, DFC and PC dynamics are exactly the same. However, in a three-layer linear network, the hidden layers' activity dynamics are not identical for the PC and DFC cases. PC dynamics follow:
\begin{equation}
\begin{aligned}
\tau \dot{\vv{v}}_{t}^{1} &= -\left(\vv{v}_{t}^{1} - \mm{W}_{t}^{1} \vv{x}\right) + \left(\mm{W}_{t}^{2}\right)^{T} \left[\vv{v}_{t}^{2} - \mm{W}_{t}^{2} \vv{v}_{t}^{1}\right] \\
\tau \dot{\vv{v}}_{t}^{2} &= -\left(\vv{v}_{t}^{2} - \mm{W}_{t}^{2} \vv{v}_{t}^{1}\right) + \left(\mm{W}_{t}^{3}\right)^{T} \left[\vv{r}^{*} - \mm{W}_{t}^{3} \vv{v}_{t}^{2}\right]
\end{aligned}
\label{eq:first-order-pc}
\end{equation}
In contrast, DFC dynamics evolve according to:
\begin{equation}
\begin{aligned}
\tau \dot{\vv{v}}_{t}^{1} &= -\left(\vv{v}_{t}^{1} - \mm{W}_{t}^{1} \vv{x}\right) + \left(\mm{W}_{t}^{2}\right)^{T} \left(\mm{W}_{t}^{3}\right)^{T} \left[\vv{r}^{*} - \mm{W}_{t}^{3} \vv{v}_{t}^{2}\right] \\
\tau \dot{\vv{v}}_{t}^{2} &= -\left(\vv{v}_{t}^{2} - \mm{W}_{t}^{2} \vv{v}_{t}^{1}\right) + \left(\mm{W}_{t}^{3}\right)^{T} \left[\vv{r}^{*} - \mm{W}_{t}^{3} \vv{v}_{t}^{2}\right]
\end{aligned}
\label{eq:first-order-dfc}
\end{equation}
We show in \Cref{pc-vs-dfc-odes} that the last layer's dynamics ($\vv{v}_{t}^{2}$) can be described by second-order ordinary differential equations for both the DFC and PC cases, i.e.,
\begin{align}
\tau^{2} \ddot{\vv{v}}_{t}^{2^{\text{PC}}} + \tau  \mm{A}_{\text{PC}} \dot{\vv{v}}_{t}^{2^{\text{PC}}} +  \mm{B} \vv{v}_{t}^{2^{\text{PC}}} &= \mm{C} \label{eq:second-order-pc} \\
\tau^{2} \ddot{\vv{v}}_{t}^{2^{\text{DFC}}} + \tau  \mm{A}_{\text{DFC}} \dot{\vv{v}}_{t}^{2^{\text{DFC}}} +  \mm{B} \vv{v}_{t}^{2^{\text{DFC}}} &= \mm{C} \label{eq:second-order-dfc}
\end{align}
where \eqref{eq:second-order-pc} follows from \eqref{eq:first-order-pc} and \eqref{eq:second-order-dfc} from \eqref{eq:first-order-dfc}.
Notably, the only coefficients that differ between these two systems are the damping coefficients, where $\mm{A}_{\text{PC}} - \mm{A}_{\text{DFC}} = \mm{W}_{t}^{2} \left(\mm{W}_{t}^{2}\right)^{T}$. This suggests that after being perturbed from their initial fixed point, both systems converge to the same steady state given by $\mm{B} \vv{v}_{*}^{2}=\mm{C}$. However, the transient dynamics will differ, with PC dynamics being more damped (as $\mm{W}_{t}^{2} \left(\mm{W}_{t}^{2}\right)^{T} \succeq 0$).

\begin{figure}[!ht]
\centering
\includegraphics[width=0.9\linewidth]{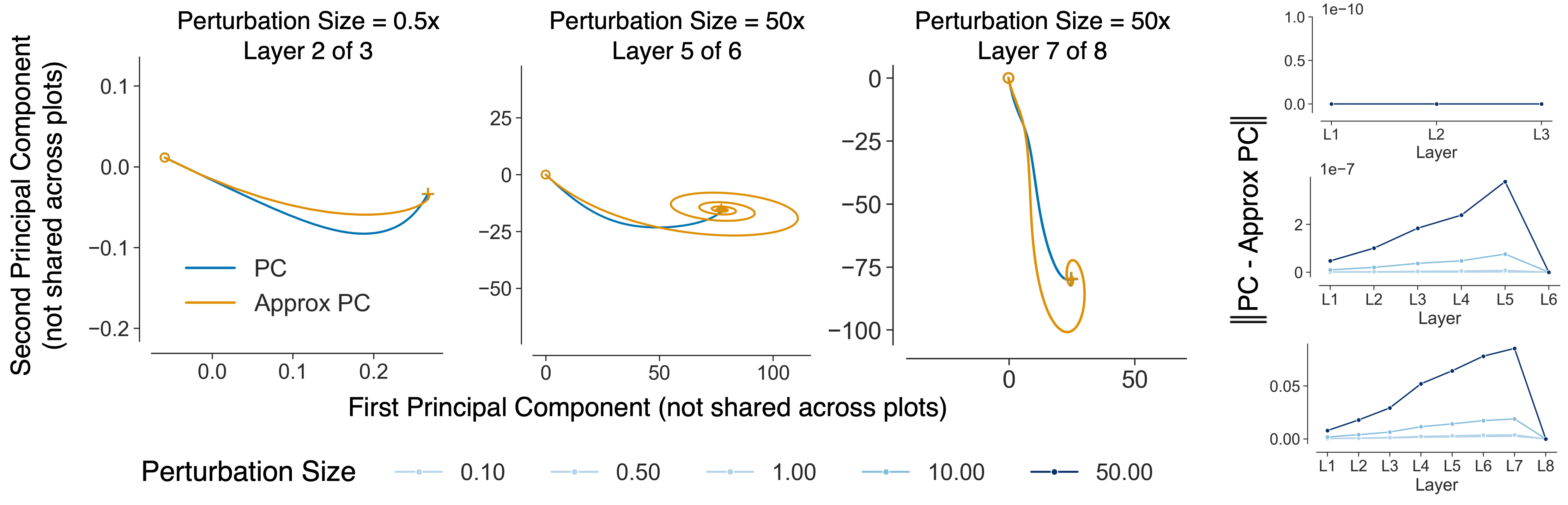}
\caption{Differences in dynamics of PC and our approximation in their first two principal components. The clamped target for all plots is the same noise perturbation direction. See \Cref{experiment-details} for details. \textbf{(Left)} The potential trajectories for individual layers across networks of increasing depth. The three-layer network converges to the same endpoint as PC, while the six- and eight-layer networks converge to different values with $L_{2}$-error on the order of $10^{-7}$ and $10^{-2}$, respectively. \textbf{(Right)} $L_{2}$-norm of the difference in final potentials across layers and perturbation sizes, for the three networks on the left.}
\label{fig:linear-comparison}
\end{figure}

We confirm these results empirically in \Cref{fig:linear-comparison}. Given a network and input, we ensure it is at a fixed point by running PC dynamics without output clamping until convergence. Then, from the converged fixed point, we perturb the system by clamping the output to a target a known distance away (reported as relative ``perturbation size''). For a three-layer network, as seen in \Cref{fig:linear-comparison} (left), both PC and our approximation converge to the same point, though they have different dynamics. As we increase the number of layers in the network to eight, the dynamics become oscillatory and converge to different values (\Cref{fig:linear-comparison}).

\begin{figure}[!ht]
\centering
\begin{minipage}{0.4\textwidth}
    \centering
    \includegraphics[width=\linewidth]{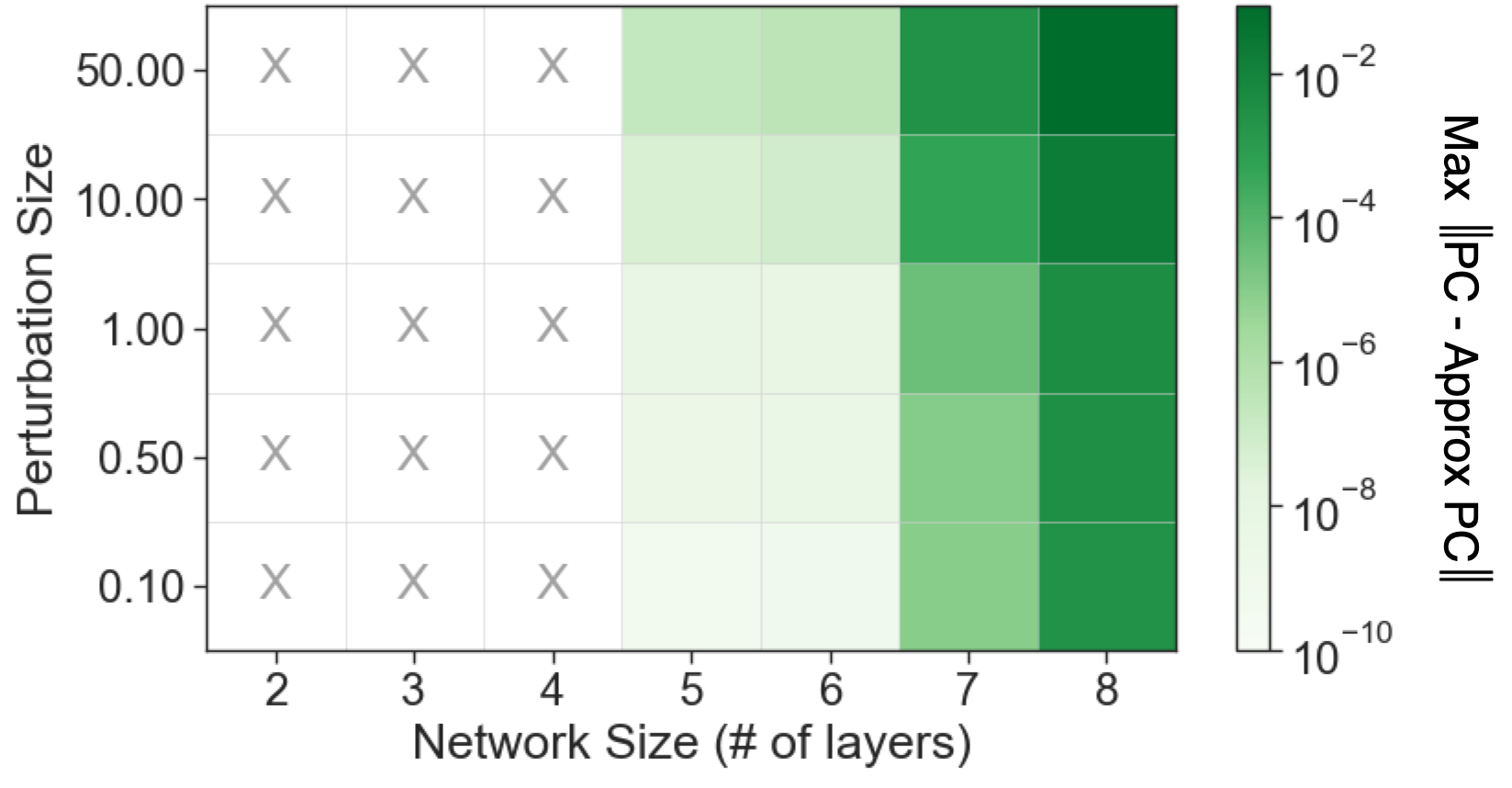}
\end{minipage}\hfill
\begin{minipage}{0.55\textwidth}
    \caption{Maximum $L_{2}$-norm error across layers in networks of varying depth for different perturbation sizes. ``X'' indicates that the error is $\leq 10^{-10}$. See \Cref{experiment-details} for details.}
    \label{fig:linear-summary}
    \vspace{5em}
\end{minipage}
\end{figure}

Looking across networks of different depth and perturbations of varying size, we observe that both perturbation magnitude and depth increase the worst-case error (\Cref{fig:linear-summary}). Notably, the error is numerically zero for networks with two or three layers, as predicted by our theory. Empirically, while this appears to extend to four-layer networks, deeper networks exhibit persistent residual error between PC and Approx PC, suggesting that the approximation may not be reliable away from fixed points. In the following section, we examine this residual error more generally for nonlinear networks and discuss techniques to mitigate it, allowing our framework to remain applicable beyond local regimes.

\subsection{Approximation characterization in nonlinear networks}

\begin{figure}[!ht]
\centering
\includegraphics[width=0.95\linewidth]{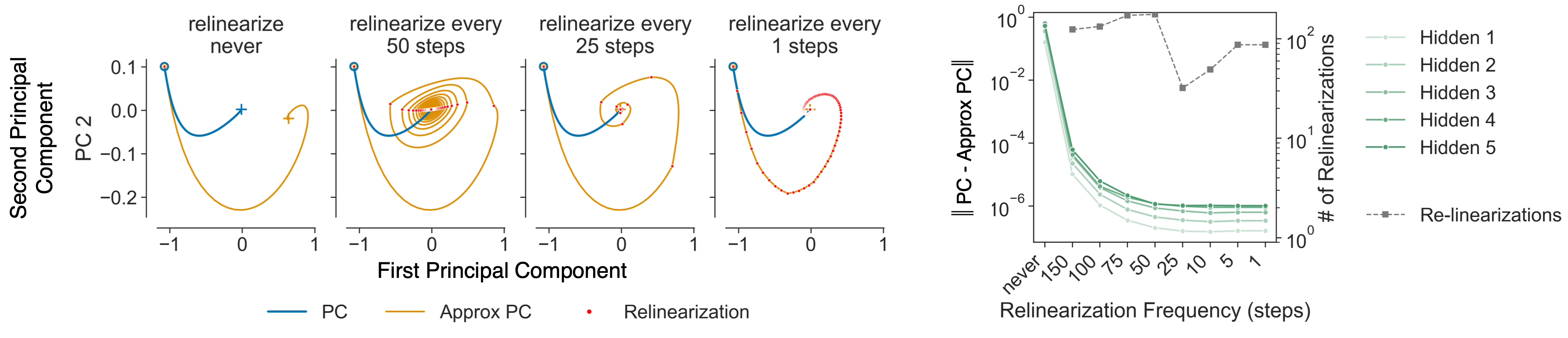}
\caption{Differences in dynamics of PC and Approx PC in their first two principal components as the frequency of relinearization is adjusted. Never relinearizing corresponds to following the dynamics of \eqref{eq:approx-dfc-dynamics}. See \Cref{experiment-details} for details. \textbf{(Left)} Trajectories of the potentials of the fourth layer in a six-layer network with $\tanh$ activation functions. \textbf{(Right)} $L_{2}$-norm error in the endpoints of PC and Approx PC across layers decreases as relinearization frequency increases. The number of relinearizations (gray line) stays $\approx$ constant until an abrupt change around a frequency of 25 steps.}
\label{fig:relinearization-comparison}
\end{figure}

While linear networks allow us to characterize how violating the assumptions in \eqref{eq:pcn_fixedpoint-control-signal} affects the approximate dynamics, we would also like to understand how the linearization of the network Jacobian produces errors away from the fixed point. To do this, we introduce the notion of relinearization time points---a set of time points, $\mathcal{T}$, that denote when to recompute the Jacobian of the network. Specifically, we modify \eqref{eq:approx-dfc-dynamics} to
\begin{equation}
\begin{aligned}
\tau_{v} \deriv{\vv{v}_{t}^{\ell}}{t} &= -\vv{v}_{t}^{\ell} + \mm{W}_{t}^{\ell} \vv{r}_{t}^{\ell - 1} + \vv{u}_{t}^{\ell} \\
\text{where} & \quad \vv{u}_{t}^{\ell} = \left(\prod_{k = \ell + 1}^{L} \mm{J}_{t'}^{k}\right)^{\top} \vv{e}_{t}^{L} \text{ and } t' = \arg \min_{\substack{t'' \in \mathcal{T} \\ t'' < t}} t - t''
\end{aligned}
\label{eq:approx-dfc-relinearize-dynamics}
\end{equation}
In practice, $\mathcal{T}$ is defined based on a relinearization frequency, i.e., the Jacobians are recomputed every $N$ steps. We examine the effect of relinearization frequency on the network dynamics for a nonlinear network in \Cref{fig:relinearization-comparison}. Without relinearization, the network fails to converge to the same final potentials as PC. With infrequent relinearization, Approx PC approaches the same endpoint as PC, but the dynamics overshoot, requiring more total integration steps and relinearizations (on the order of $10^{2}$) to converge. This persists until a relinearization frequency of 25 steps, at which point the dynamics are still underdamped compared to PC, but overshoot no longer occurs. As a result, the system converges faster and requires an order of magnitude fewer relinearizations. Increasing the relinearization frequency beyond this point does not fundamentally change the dynamics, and only increases the total number of relinearizations without meaningfully reducing the final endpoint difference between PC and Approx PC.

\begin{figure}[!ht]
\centering
\includegraphics[width=0.95\linewidth]{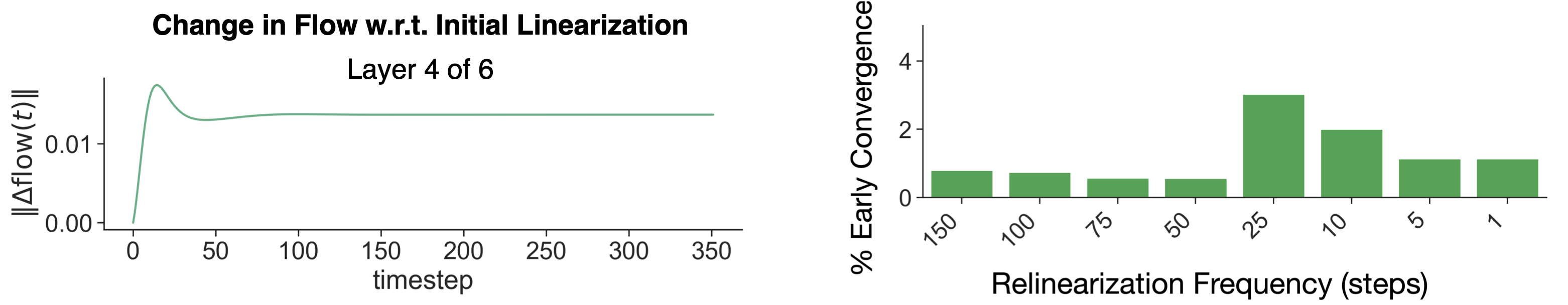}
\caption{\textbf{(Left)} The $L_{2}$-norm of the change in flow w.r.t. the initial linearized flow moving along the Approx PC trajectory without relinearization in \Cref{fig:relinearization-comparison}. \textbf{(Right)} Percentage of segments between linearizations where the dynamics of Approx PC converge (i.e., the speed of dynamics $\leq 10^{-8}$) earlier than the relinearization frequency. See \Cref{experiment-details} for details.}
\label{fig:relinearization-speed}
\end{figure}

We note that there is nothing fundamental about 25 steps. Instead, given the initial fixed point of the system, its nonlinear dynamics, and a particular perturbation, 25 steps represents the distance the system is allowed to move away from the fixed point without incurring significant error due to linearization. We measure this in \Cref{fig:relinearization-speed} (left) as the difference in flow w.r.t. the initial linearized flow as we move along the Approx PC trajectory without relinearization. Essentially, this measures at each timestep how much the dynamics would change if we relinearize. The peak at 25 steps indicates the moment at which we would achieve the largest error correction. By relinearizing at this point, we move more efficiently toward the PC endpoint and quickly benefit from the slow dynamics of the nonlinear system around the PC endpoint (\Cref{fig:relinearization-speed}, right). Thus, we conclude that relinearization can reduce the residual error of Approx PC \emph{efficiently}, provided that we do not move ``too far'' before reorienting the dynamics.

\section{Resolving residual errors with nudging}

In the previous section, we showed that Approx PC without relinearization exhibits residual errors compared to PC for all but the simplest linear models. While relinearization addresses this issue, its timing greatly affects dynamical efficiency, and this depends on the network, the current fixed point, and the target. In a realistic training scenario, it is not clear how these three factors interact.

To address this, we introduce a multi-stage algorithm (\Cref{alg:approx-pc-nudges}) in which the trajectory toward a target is segmented into multiple smaller subgoals. Each subgoal is a ``nudge,'' i.e., a partial step toward the target, computed as a fraction of the linear path between the current output and the target (see step 2(a) of \Cref{alg:approx-pc-nudges}). Importantly, after each nudge, the dynamics are allowed to relax to a new fixed point, after which a new subgoal is computed and the next nudge is initiated.

\begin{figure}[!ht]
\centering
\includegraphics[width=0.95\linewidth]{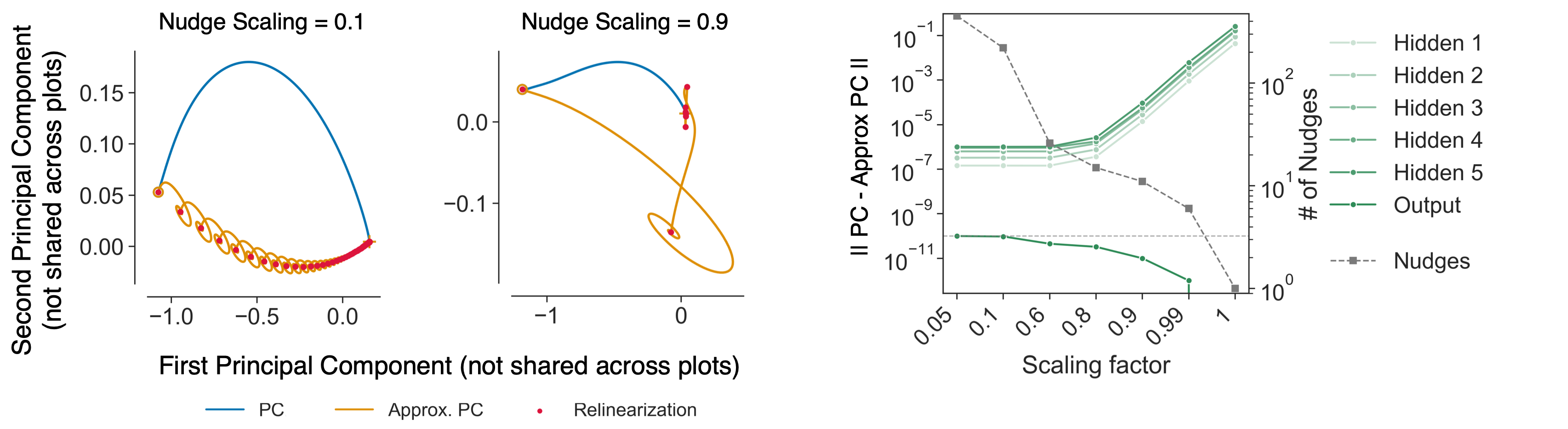}
\caption{Differences in dynamics of PC and Approx PC in their first two principal components for different nudge scalings. A scaling factor of one corresponds to following the dynamics of \eqref{eq:approx-dfc-dynamics}. The perturbation size is one. See \Cref{experiment-details} for details. \textbf{(Left)} Trajectories of the potentials of the fourth layer in a six-layer network with $\tanh$ activation functions for two different nudge scaling factors. Nudges are recomputed at relinearization points. \textbf{(Right)} $L_{2}$-norm error in the endpoints of PC and Approx PC across layers decreases with smaller scaling factors. The number of nudges (gray line) consistently decreases with larger scaling factors. The horizontal dashed line indicates the convergence threshold for nudging.}
\label{fig:nudge-comparison}
\end{figure}

\Cref{fig:nudge-comparison} shows the resulting trajectories when we apply \Cref{alg:approx-pc-nudges}. If the nudge scaling factor, $\gamma$, is small, the network follows a frequent relinearization-like trajectory (albeit with small oscillations to convergence between nudges). With large scaling factors, we observe a new behavior where the trajectory initially deviates from the PC endpoint significantly (similar to never relinearizing); however, since the network is allowed to relax to a fixed point, the next nudge reorients the trajectory toward the PC endpoint. This is in contrast to infrequent relinearization, which keeps overshooting the PC endpoint. As expected, the final difference between the endpoints of PC and Approx PC with nudges is small (of the same order as with relinearization). Notably, this applies to large scaling factors as well, where the total number of nudges is small. In this way, nudging preserves the low residual error of relinearization without consuming a significant number of additional integration steps.

\begin{algorithm}[!ht]
\caption{Approx PC with nudges}
\label{alg:approx-pc-nudges}
\begin{enumerate}
\item \textbf{Requires:} a network of $L$ layers at fixed point $\vv{v}_{*}$, input $\vv{x}$, target $\vv{y}$, nudge scaling factor $\gamma \in \left(0, 1\right]$, nudge convergence threshold $\epsilon > 0$, dynamics convergence threshold $\delta > 0$
\item \textbf{while} $\|\vv{v}_{*}^L - \vv{y}\| > \epsilon$ \textbf{do}
\begin{enumerate}
\item \underline{compute nudge:} $\vv{r}_* \leftarrow \vv{v}_*^L + \gamma (\vv{y} - \vv{v}_*^L)$
\item \underline{clamp network:} $\vv{r}^L \leftarrow \vv{r}_*$
\item \underline{compute Jacobian:} $\mm{J}^\ell \leftarrow$ layer-wise Jacobian at current timestep for input $\vv{x}$ for all $\ell$
\item \underline{run dynamics until convergence:} \textbf{do}
\begin{enumerate}
\item $\vv{e}^L \leftarrow \vv{v}^L - \mm{W}^L \vv{r}^{L - 1}$
\item $\vv{u}^\ell \leftarrow (\prod_{k = \ell + 1}^L \mm{J}^k)^\top \vv{e}^L$ for all $\ell$
\item $\vv{v}_\text{prev}^\ell \leftarrow \vv{v}^\ell$
\item $\vv{v}^\ell \leftarrow \vv{v}^\ell + \tau_v^{-1} (-\vv{v}^\ell + \mm{W}^\ell \vv{r}^{\ell - 1} + \vv{u}^\ell)$ for all $\ell$
\item $\mm{W}^\ell \leftarrow \mm{W}^\ell + \tau_w^{-1} [\vv{u}^\ell (\vv{r}^{\ell - 1})^\top]$ for all $\ell$
\end{enumerate}
\par \textbf{while} $\|\vv{v}^{\ell} - \vv{v}_{\text{prev}}^{\ell}\| > \delta$
\item \underline{update fixed point:} $\vv{v}_* \leftarrow \vv{v}$
\end{enumerate}
\end{enumerate}
\end{algorithm}

\section{Approx PC with learning}

To evaluate whether our Approx PC algorithm remains faithful to PC in practical settings with learning, we implemented and examined both approaches on standard supervised image classification tasks (see \Cref{experiment-details} for details). We first tested whether networks trained with Approx PC converge to solutions that are compatible with predictive coding. To this end, we trained a network using Approx PC and then directly transplanted the learned weights into a PC model without further training for evaluation. \textbf{Approx PC networks achieved a final test accuracy of $97.32 \pm 0.12$\% on MNIST across five seeds, and the corresponding PC networks achieved an evaluated test accuracy of $97.47 \pm 0.05$\% across the same seeds.} This confirms that Approx PC learns solutions compatible with PC dynamics.

\begin{figure}[!ht]
\centering
\includegraphics[width=\linewidth]{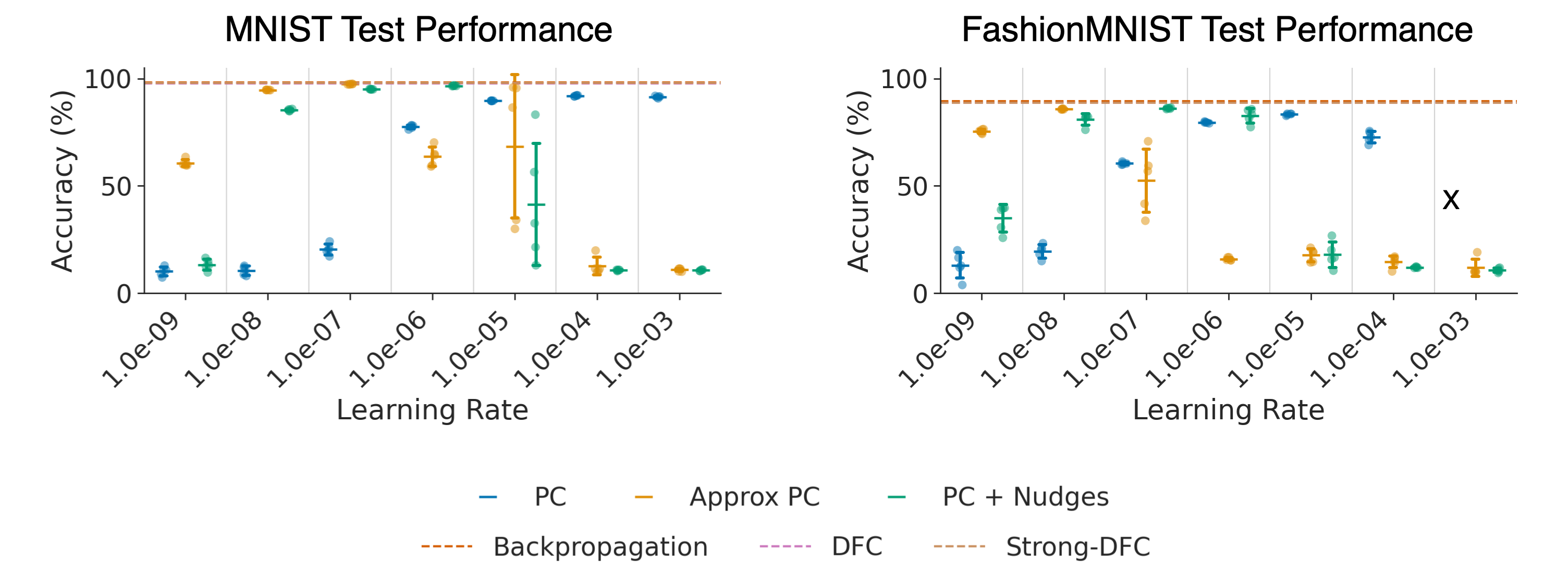}
\caption{Final test accuracy on MNIST (left) and FashionMNIST (right) across learning rates for PC, Approx PC, and PC with nudges (mean and standard deviation across five seeds). Dashed lines show previously reported test accuracies for DFC \cite{MeulemansCreditAssignmentNeural2021},  Strong-DFC \cite{MeulemansMinimizingControlCredit2022}, and standard backprop (via \cite{MeulemansMinimizingControlCredit2022}). Some configurations did not learn and are excluded. ``X'' denotes configurations with no successful runs. $\text{Learning rate} := \tau_w^{-1}$.}
\label{mnist}
\end{figure}

For both MNIST and FashionMNIST, Approx PC reaches accuracy similar to or higher than that of standard PC at their respective best learning rates (\Cref{mnist}). Interestingly, Approx PC also seems to learn faster per integration step (\Cref{learning-speed}). This could be due to the fact that Approx PC propagates a centralized error, allowing layers to converge concurrently, while PC uses distributed error signals that must propagate across the network one layer at a time. However, Approx PC also uses multiple nudges per sample, so the improvement may instead come from breaking each learning step into intermediate subgoals. To isolate this effect, we implemented PC with nudges, where the network is trained using intermediate nudges as clamped outputs but with standard PC dynamics. We found that adding nudges to PC recovers most of the learning performance and efficiency enjoyed by Approx PC (\Cref{mnist,learning-speed}). This suggests that most of the gain over standard PC comes from nudging rather than from the centralized controller. However, for the same number of samples seen, Approx PC and PC with nudges require more total integration steps than standard PC. Finally, hyperparameter sweeps on FashionMNIST (\Cref{tab:results}) show that both methods degrade similarly with network depth, but differ in their sensitivity to other hyperparameters. In particular, Approx PC is less sensitive to the neural time constant and dynamics convergence tolerance, but has a slightly narrower range of effective learning rates. 

\section{Discussion}
\label{sec:discussion}

We have shown that predictive coding and deep feedback control, two prominent frameworks for biologically plausible deep learning, can be understood within a common dynamical systems perspective. In particular, DFC provides a local approximation to PC dynamics near fixed points, revealing that these seemingly distinct approaches differ primarily in their transient dynamics rather than their steady-state behavior.

\textbf{Implications for machine learning.} We have shown that Approx PC reaches approximately the same fixed points as PC, while differing in its transient dynamics. Moreover, weights learned under Approx PC can be transplanted into PC networks for inference. These results suggest that part of the slow inference attributed to PC may stem from its particular dynamics rather than from energy-based learning itself. In PC, errors reach early layers only by propagating one layer at a time. Separately, nudging accounted for most of the accuracy gain over standard PC in our experiments. This may partly reflect the extra weight updates that nudging entails. Another possibility is that clamping the network's output directly to the label may cause it to reach a different equilibrium than would be reached by moving toward the label more gradually. Equilibrium propagation makes a similar argument for weak clamping, where small nudges keep the network's new equilibrium close to its original one~\citep{ScellierEquilibriumPropagationBridging2017}. Whether moving the target gradually benefits energy-based learning more broadly remains to be tested. 

\textbf{Implications for biological interpretation.} Control-theoretic and predictive coding frameworks have often been proposed as alternative accounts of credit assignment in the brain, distinguished by whether error signals are computed locally at each layer or generated centrally and broadcast through feedback pathways. Our results indicate that, near fixed points, PC dynamics can be expressed as an equivalent control strategy driven by output errors, making the two frameworks difficult to tell apart based on steady-state activity alone. Our theory instead suggests that distinguishing these mechanisms in the brain requires an analysis of transient activity. In particular, perturbations in one layer of a PC network immediately affect the preceding layer through local errors, whereas under DFC, preceding layers respond only after the perturbation has reached the output. We therefore suggest experiments that perturb one stage of a processing hierarchy and compare the timing of responses at earlier stages and at the output.

\textbf{Limitations and future work.} The primary purpose of our experiments was to validate predictions arising from the theoretical connection made between PC and DFC. Therefore, we focused on controlled experiments in fully connected networks, where making sense of activity and learning dynamics is more tractable. While we demonstrated the practical feasibility of Approx PC through MNIST and FashionMNIST experiments, future work should expand our approach to neural networks with more parameters and different architectures (e.g., convolutional networks), and assess whether Approx PC can be competitive with established training methods at scale. The appropriate nudge size and convergence thresholds also depend on the network and data, making them difficult to set in advance. In practice, we found the number of steps per nudge to be a useful indicator of convergence behavior, which future work could formalize into an adaptive nudging scheme. Beyond machine learning applications, an implementation of Approx PC in biology is unlikely to use the exact network Jacobian. Instead, one may consider slow-varying estimates of the Jacobians or arbitrary, structured feedback weights, whose exploration we leave to future work. Finally, we hope our work can inspire extensions that explore the use of control-theoretic improvements to PC training, such as PID control.


\begin{ack}
This work was supported by the Cold Spring Harbor Laboratory Fellows Program and Pershing Square Foundation. The authors declare no competing interests.
\end{ack}

\bibliographystyle{plainnat}
\bibliography{ref}

\newpage
\appendix
\crefalias{section}{appendix}
\crefalias{subsection}{subappendix}

\section{Rewriting PC activity dynamics}
\label{pc_activity_deriv}

Starting from PC potential dynamics given by \eqref{eq:pc-potential-dynamics}, we rewrite the dynamics in a form compatible with DFC. To begin, we will focus on layers $\ell < L$, where the inference and learning activity is governed by the same set of equations. Let us briefly review the derivation of \eqref{eq:pc-potential-dynamics}. The gradient of the total energy is

\begin{equation}
\nabla_{\vv{v}_{t}^{\ell}} E_{t} = \frac{1}{2} \nabla_{\vv{v}_{t}^{\ell}} \|\vv{e}_{t}^{\ell}\|_{2}^{2} + \frac{1}{2} \nabla_{\vv{v}_{t}^{\ell}} \|\vv{e}_{t}^{\ell + 1}\|_{2}^{2}
\end{equation}

The other terms in \eqref{eq:pc-energy} do not depend on $\vv{v}^{\ell}$, so their derivatives are zero. Next, we can substitute in the layer-wise errors and proceed as follows

\begin{equation}
\begin{aligned}
\nabla_{\vv{v}_{t}^{\ell}} E_{t} &= \frac{1}{2} \nabla_{\vv{v}_{t}^{\ell}} \|\vv{v}_{t}^{\ell} - \mm{W}_{t}^{\ell} \vv{r}_{t}^{\ell - 1}\|_{2}^{2} + \frac{1}{2} \nabla_{\vv{v}_{t}^{\ell}} \|\vv{v}_{t}^{\ell+1} - \mm{W}_{t}^{\ell+1} \vv{r}_{t}^{\ell}\|_{2}^{2} \\
&= \underbrace{\vv{v}_{t}^{\ell} - \mm{W}_{t}^{\ell} \vv{r}_{t}^{\ell-1}}_{\vv{e}_{t}^{\ell}} - \underbrace{\phi'\left(\vv{v}_{t}^{\ell}\right) \odot \left(\mm{W}_{t}^{\ell + 1}\right)^{\top}}_{\left(\mm{J}_{t}^{\ell+1}\right)^{\top}} \underbrace{\left(\vv{v}_{t}^{\ell+1} - \mm{W}_{t}^{\ell+1} \vv{r}_{t}^{\ell}\right)}_{\vv{e}_{t}^{\ell+1}}
\end{aligned}
\end{equation}

Finally, the potential evolves according to the negative gradient of the energy, resulting in the following dynamics

\begin{equation}
\tau_{v} \deriv{\vv{v}_{t}^{\ell}}{t} = -\vv{v}_{t}^{\ell} + \mm{W}_{t}^{\ell} \vv{r}_{t}^{\ell - 1} + \left(\mm{J}_{t}^{\ell + 1}\right)^{\top} \vv{e}_{t}^{\ell + 1}
\end{equation}

This matches the form of DFC dynamics in \eqref{eq:dfc-dynamics}.

\section{Schematics for PC and DFC style controllers}

\begin{figure}[!ht]
\centering
\includegraphics[width=0.9\linewidth]{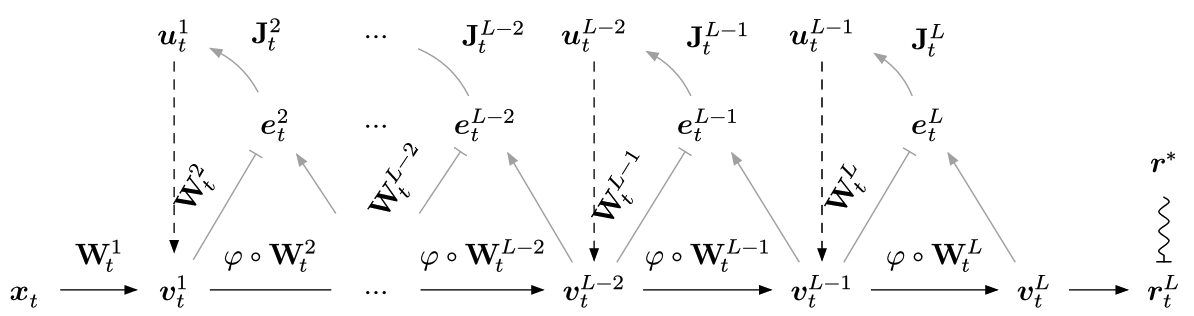}
\caption{A schematic of the ``control system'' view of predictive coding. The output is clamped to an arbitrary set point, $\vv{r}^{*}$, and each layer's activity is modulated via a feedback controller driven by the subsequent layer's error. Edges meet to indicate addition; edges with flat ends denote negative (inhibitory) signals, gray edges highlight control signals, and wavy edges denote clamping.}
\label{fig:pc-controller}
\end{figure}

\begin{figure}[!ht]
\centering
\includegraphics[width=0.9\linewidth]{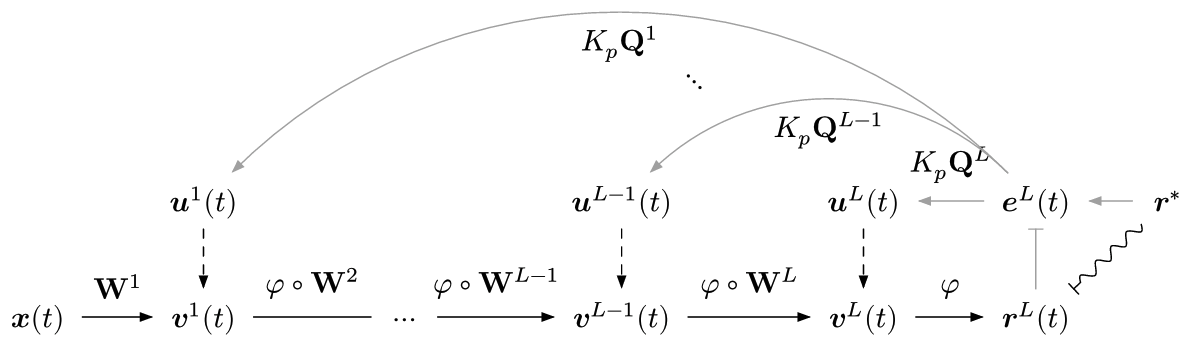}
\caption{A schematic of DFC-style control. The output is clamped to an arbitrary set point, $\vv{r}^*$, and each layer's activity is modulated via a feedback controller driven by the output error.}
\label{fig:dfc-controller}
\end{figure}

\section{Derivation of PC and Approx PC exact dynamics in linear networks}
\label{pc-vs-dfc-odes}

In this appendix we derive the ordinary differential equations (ODEs) governing layer-wise representation for both PC and Approx PC. We consider a three-layer network ($x,\vv{v}^{1},\vv{v}^{2},\vv{v}^{3}$), where the last layer $\vv{v}^{3}$ is clamped to the training output, $\vv{r}^{*}$. We also consider linear activations ($\phi(\vv{v^{\ell}}) = \vv{v^{\ell}}$, $\phi'\left(\vv{v}^{\ell}\right) = \mathbf{1}$). For both PCN and DFC approximation dynamics, the goal is to yield a closed-form ODE purely in $\vv{v^{\ell}}$. We then define the residual between the two systems and compare them.

Throughout, we consider $\vv{v}^{1} \in \mathbb{R}^{d_{1}}$, $\vv{v}^{2} \in \mathbb{R}^{d_{2}}$, input $\vv{x} \in \mathbb{R}^{d_{0}}$, and target $\vv{r}^{*} \in \mathbb{R}^{d_{3}}$. Weight matrices are $\mm{W}^{1} \in \mathbb{R}^{d_{1} \times d_{0}}$, $\mm{W}^{2} \in \mathbb{R}^{d_{2} \times d_{1}}$, $\mm{W}^{3} \in \mathbb{R}^{d_{3} \times d_{2}}$. We leave off the subscript $t$ for clarity. The time constant $\tau_v$ is shared by all layers.

\subsection{PC activity dynamics}

Recall \eqref{eq:pc-potential-dynamics}:
\begin{align}
\tau \dot{\vv{v}}^{1} &= -\left(\vv{v}^{1} - \mm{W}^{1} \vv{x}\right) + {(\mm{W}^{2})}^{T} \left[\vv{v}^{2} - \mm{W}^{2} \vv{v}^{1}\right] \label{pcn1} \\
\tau \dot{\vv{v}}_{2} &= -\left(\vv{v}^{2} - \mm{W}^{2} \vv{v}^{1}\right) + {(\mm{W}^{3}})^{T} \left[\vv{r}^{*} - \mm{W}^{3} \vv{v}^{2}\right] \label{pcn2}
\end{align}
Expanding \eqref{pcn1} and \eqref{pcn2} and collecting terms in $v_{1}$ and $v_{2}$:
\begin{align}
\tau \dot{\vv{v}}^{1} &= -\left(\mm{I} + {(\mm{W}^{2})}^{T} \mm{W}^{2}\right) \vv{v}^{1} + {(\mm{W}^{2})}^{T} \vv{v}^{2} + \mm{W}^{1}\vv{x} \label{pcn1exp} \\
\tau \dot{\vv{v}}^{2} &= -\left(\mm{I} + {(\mm{W}^{3})}^{T} \mm{W}^{3}\right) \vv{v}^{2} + \mm{W}^{2} \vv{v}^{1} + {(\mm{W}^{3})}^{T} \vv{r}^{*} \label{pcn2exp}
\end{align}
From \eqref{pcn2exp} we isolate the $\vv{v}^{1}$ -dependent term:
\begin{equation}
\mm{W}^{2} \vv{v}^{1} = \tau \dot{\vv{v}}^2 + \left(\mm{I} + {(\mm{W}^{3})}^{T} \mm{W}^{3}\right) \vv{v}^{2} - {(\mm{W}^{3})}^{T} \vv{r}^{*}
\label{pcn_v1_left}
\end{equation}
Differentiating \eqref{pcn2exp} with respect to time gives:
\begin{equation}
\tau \ddot{\vv{v}}^2 = - \left(\mm{I} + {(\mm{W}^{3})}^{T} \mm{W}^{3}\right) \dot{\vv{v}}^2 + \mm{W}^{2} \dot{\vv{v}}^1
\end{equation}
Substituting $\dot{\vv{v}}^1$ from \eqref{pcn1exp} gives:
\begin{equation}
\tau^{2} \ddot{\vv{v}}^2 = - \tau \left(\mm{I} + {(\mm{W}^{3})}^{T} \mm{W}^{3}\right) \dot{\vv{v}}^2 + \mm{W}^{2} \left[- \left(\mm{I} + {(\mm{W}^{2})}^{T} \mm{W}^{2}\right) \vv{v}^{1} + {(\mm{W}^{2})}^{T} \vv{v}^{2} + \mm{W}^{1} \vv{x}\right]
\label{sec_degree_v2}
\end{equation}
Note that $\vv{v}^1$ coefficient can be written as:
\begin{equation}
- \mm{W}^{2} \left(\mm{I} + {(\mm{W}^{2})}^{T} \mm{W}^{2}\right) \vv{v}^{1} = - \left(\mm{I} + \mm{W}^{2} {(\mm{W}^{2})}^{T}\right) \left(\mm{W}^{2} \vv{v}^{1}\right)
\end{equation}
and we can replace $\mm{W}^{2} \vv{v}^{1}$ from \eqref{pcn_v1_left}, and write \eqref{sec_degree_v2} as:
\begin{equation}
\begin{aligned}
\tau^{2} \ddot{\vv{v}}^2 = & - \tau \left(\mm{I} + {(\mm{W}^{3})}^{T} \mm{W}^{3}\right) \dot{\vv{v}}^2 \\
& - \left(\mm{I} + \mm{W}^{2} {(\mm{W}^{2})}^{T}\right) \left[\tau \dot{\vv{v}}^2 + \left(\mm{I} + {(\mm{W}^{3})}^{T} \mm{W}^{3}\right) \vv{v}^{2} - {(\mm{W}^{3})}^{T} \vv{r}^{*}\right] \\
& + \mm{W}^{2} {(\mm{W}^{2})}^{T} \vv{v}^{2} + \mm{W}^{2} \mm{W}^{1} \vv{x}
\end{aligned}
\end{equation}
Grouping terms, the final PCN second-order ODE for $\vv{v}_{2}$ can therefore be expressed as:
\begin{equation}
\begin{aligned}
\tau^{2} \ddot{\vv{v}}^2 + \tau \underbrace{\left(2 \mm{I} + \mm{W}^{2} {(\mm{W}^{2})}^{T} + {(\mm{W}^{3})}^{T} \mm{W}^{3}\right)}_{A_{\text{PCN}}} \dot{\vv{v}}^2 + \underbrace{\left(\mm{I} + {(\mm{W}^{3})}^{T} \mm{W}^{3} + \mm{W}^{2} {(\mm{W}^{2})}^{T} {(\mm{W}^{3})}^{T} \mm{W}^{3}\right)}_{B_{\text{PCN}}} \vv{v}^{2} \\ = \underbrace{\mm{W}^{2} \mm{W}^{1} \vv{x} + \left(\mm{I} + \mm{W}^{2} {(\mm{W}^{2})}^{T}\right) {(\mm{W}^{3})}^{T} \vv{r}^{*}}_{C_{\text{PCN}}}
\end{aligned}
\label{pcn_final}
\end{equation}

\subsection{Approx PC activity dynamics}

In Approx PC, $\mm{Q}^{\ell}$ is approximated by the product of the Jacobians of the following layers, that is:
\begin{align}
\tau \dot{\vv{v}}^{1} &= -\left(\vv{v}^{1} - \mm{W}^{1} \vv{x}\right) + (\mm{W}^{2})^{T} (\mm{W}^{3})^{T} \left[\vv{r}^{*} - \mm{W}^{3} \vv{v}^{2}\right] \label{dfc1} \\
\tau \dot{\vv{v}}^{2} &= -\left(\vv{v}^{2} - \mm{W}^{2} \vv{v}^{1}\right) + (\mm{W}^{3})^{T} \left[\vv{r}^{*} - \mm{W}^{3} \vv{v}^{2}\right] \label{dfc2}
\end{align}
Expanding \eqref{dfc1} and \eqref{dfc2}:
\begin{align}
\tau \dot{\vv{v}}^{1} &= -\vv{v}^{1} + \mm{W}^{1} \vv{x} + (\mm{W}^{2})^{T} (\mm{W}^{3})^{T} \vv{r}^{*} - (\mm{W}^{2})^{T} (\mm{W}^{3})^{T} \mm{W}^{3} \vv{v}^{2} \label{dfc1exp} \\
\tau \dot{\vv{v}}^{2} &= -\left(\mm{I} + (\mm{W}^{3})^{T} \mm{W}^{3}\right) \vv{v}^{2} + \mm{W}^{2} \vv{v}^{1} + (\mm{W}^{3})^{T} \vv{r}^{*} \label{dfc2exp}
\end{align}
From \eqref{dfc2exp} we isolate the $\vv{v}^{1}$-dependent term:
\begin{equation}
\mm{W}^{2} \vv{v}^{1} = \tau \dot{\vv{v}}^2 + \left(\mm{I} + (\mm{W}^{3})^{T} \mm{W}^{3}\right) \vv{v}^{2} - (\mm{W}^{3})^{T} \vv{r}^{*}
\label{dfc_v2}
\end{equation}
Differentiating \eqref{dfc2exp}:
\begin{equation}
\tau^{2} \ddot{\vv{v}}^2 = - \tau \left(\mm{I} + (\mm{W}^{3})^{T} \mm{W}^{3}\right) \dot{\vv{v}}^2 + \mm{W}^{2} \left(\tau \dot{\vv{v}}^1\right)
\end{equation}
Replacing $\dot{\vv{v}}^1$ from \eqref{dfc1exp} gives:
\begin{equation}
\begin{aligned}
\tau^{2} \ddot{\vv{v}}^2 &= - \tau \left(\mm{I} + (\mm{W}^{3})^{T} \mm{W}^{3}\right) \dot{\vv{v}}^2 \\
& \quad + \mm{W}^{2} \left[- \vv{v}^{1} + \mm{W}^{1} \vv{x} + (\mm{W}^{2})^{T} (\mm{W}^{3})^{T} \vv{r}^{*} - (\mm{W}^{2})^{T} (\mm{W}^{3})^{T} \mm{W}^{3} \vv{v}^{2}\right]
\end{aligned}
\end{equation}
And finally replacing $\mm{W}^{2} \vv{v}^{1}$ from \eqref{dfc_v2} and grouping terms gives us the final DFC second-order ODE:
\begin{equation}
\begin{aligned}
\tau^{2} \ddot{\vv{v}}^2 + \tau \underbrace{\left(2 \mm{I} + (\mm{W}^{3})^{T} \mm{W}^{3}\right)}_{A_{\text{DFC}}} \dot{\vv{v}}^2 + \underbrace{\left(\mm{I} + (\mm{W}^{3})^{T} \mm{W}^{3} + \mm{W}^{2} (\mm{W}^{2})^{T} (\mm{W}^{3})^{T} \mm{W}^{3}\right)}_{B_{\text{DFC}}} \vv{v}^{2} \\ = \underbrace{\mm{W}^{2} \mm{W}^{1} \vv{x} + \left(\mm{I} + \mm{W}^{2} (\mm{W}^{2})^{T}\right) (\mm{W}^{3})^{T} \vv{r}^{*}}_{C_{\text{DFC}}}
\end{aligned}
\label{dfc_final}
\end{equation}

\subsection{Residual between PCN and DFC}

Comparing (28) and (36), the two second-order ODEs have identical stiffness ($\mathbf{B}$) and forcing ($\mathbf{C}$) terms, and differ only in their damping matrices:
\begin{equation}
\mathbf{A}_{\text{PCN}} - \mathbf{A}_{\text{DFC}} = \mathbf{W}^2(\mathbf{W}^2)^\top.
\end{equation}
Hence any fixed point of one system
is also a fixed point of the other, and the two differ only in their transient dynamics, with PC carrying an additional damping term.

\section{Experimental details}
\label{experiment-details}

\subsection{Hyperparameters}
\label{sec:hyperparams}

In this section we specify the hyperparameters used to generate the results presented in this paper.
For all experiments, the proportional control constant is $K_p=1$. For \Cref{fig:linear-comparison,fig:linear-summary,fig:relinearization-comparison,fig:relinearization-speed,fig:nudge-comparison}, the width of hidden layers is eight, the input and output dimensions are one, and the neural dynamics time constant is $\tau_v = 10$. The perturbation direction is a single unit vector sampled from $\mathcal{N}(0, \mm{I})$ and shared across comparisons; only its magnitude is varied where specified. For \Cref{fig:linear-comparison,fig:linear-summary}, the dynamics convergence tolerance is \num{1e-15}. For \Cref{fig:relinearization-comparison,fig:relinearization-speed,fig:nudge-comparison}, we used networks with five hidden layers and a dynamics convergence tolerance of \num{1e-8}. For \Cref{fig:nudge-comparison}, the nudge convergence tolerance is \num{1e-10}.
For the MNIST and FashionMNIST experiments, by default, two hidden layers of width 256 are used; the neural dynamics time constant is $\tau_v = 30$; the dynamics convergence tolerance is \num{1e-2}; the nudge convergence tolerance is \num{1e-2}; the nudge scaling factor is $\gamma = 0.1$; networks have a bias (all other experiments have no bias); and we use the Adam optimizer with weight decay of \num{1e-4}. \Cref{fig:linear-comparison,fig:linear-summary} concern linear networks (identity activation functions), while all other experiments use $\tanh$ activation functions in the hidden layers.

\subsection{Computational resources}
\label{sec:resources}

\begin{figure}[!ht]
\centering
\includegraphics[width=0.7\linewidth]{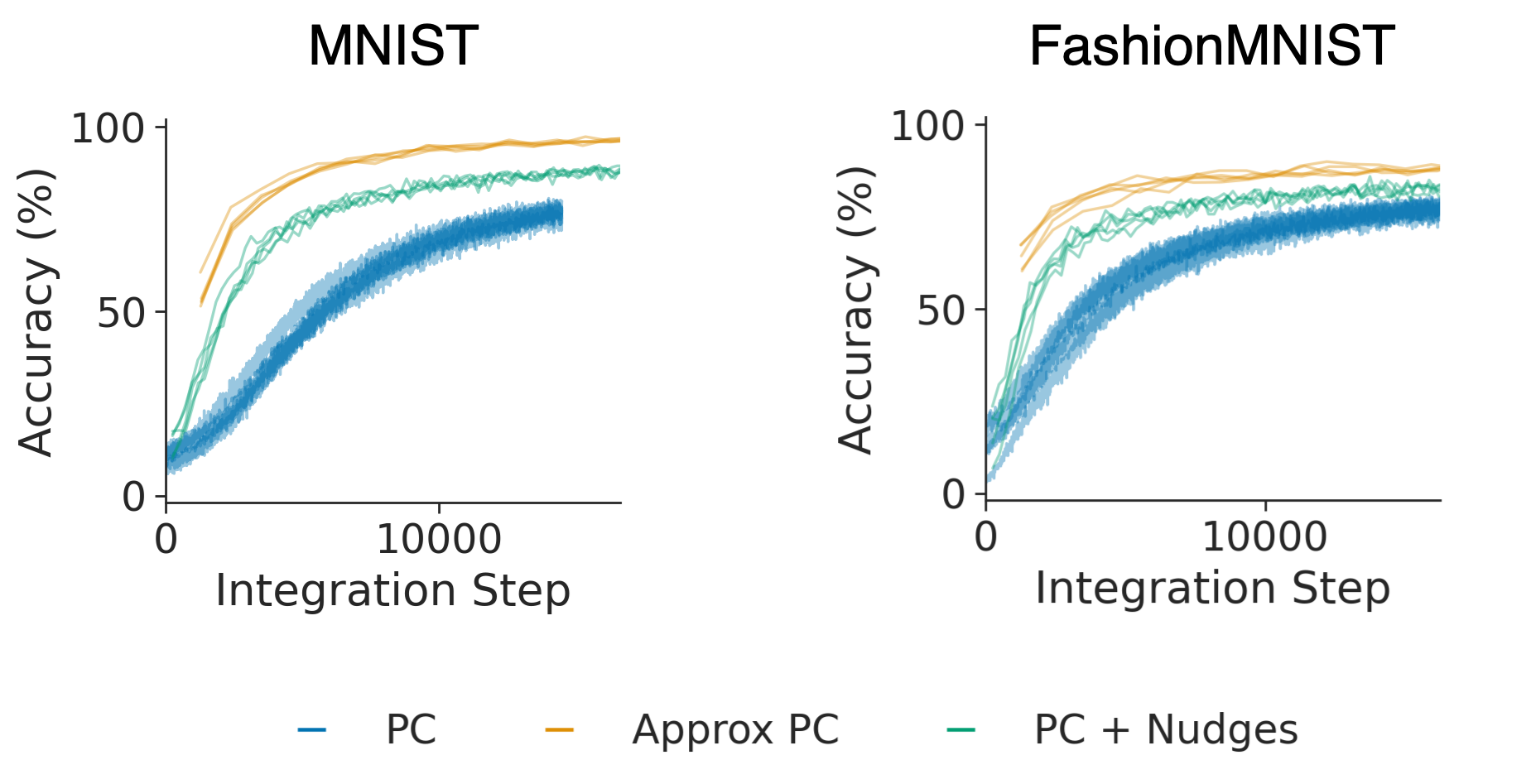}
\caption{Results on MNIST (left) and FashionMNIST (right) for PC, Approx PC, and PC with nudges across five seeds. The total number of training samples is fixed, but the number of integration steps per sample varies. Per integration step, Approx PC and PC with nudges improve in test accuracy more than PC. The first data point is recorded after the first sample (including multiple nudges). The x-axis is cropped to zoom in on the initial learning curve ramp-up where the effect is most apparent. Note that PC takes fewer total integration steps to process the same total number of samples, which is reflected in the MNIST results where the PC learning curve stops earlier than those of the nudged variants. $\text{Learning rate} := \tau_w^{-1} =$ \num{1e-6} (chosen so that the effect is easiest to visualize though this result holds across learning rates tested in the paper).}
\label{learning-speed}
\end{figure}

For \Cref{fig:linear-comparison,fig:linear-summary,fig:relinearization-comparison,fig:relinearization-speed,fig:nudge-comparison}, we used a MacBook Pro with M2 processor and 16GB of RAM. For \Cref{mnist,learning-speed} and \Cref{tab:results}, we used a single H100 GPU with 96GB of VRAM per learning rate and seed configuration.

\section{Additional experimental results}
\label{additional-results}

\begin{table}[!ht]
    \centering
    \caption{Hyperparameter sweep results for FashionMNIST. Unless a parameter is being swept, the default values are time constant = 30, dyn. conv. tol. = \num{1e-2}, nudge conv. tol. = \num{1e-2}, \# of layers = 2, and learning rate = \num{1e-5} (PC) or \num{1e-8} (Approx PC). Values are mean and standard deviation over five random seeds.}
    \label{tab:results}
    \begin{tabular}{rrrrr|rr}
    \toprule
    \textbf{LR} & \shortstack[r]{\textbf{Time} \\ \textbf{constant}} & \shortstack[r]{\textbf{Dynamics} \\ \textbf{conv. tol.}} & \shortstack[r]{\textbf{Nudge} \\ \textbf{conv. tol.}} & \shortstack[r]{\textbf{\# of} \\ \textbf{layers}} & \textbf{PC Acc. (\%)} & \shortstack[r]{\textbf{Approx} \\ \textbf{PC Acc. (\%)}} \\
    \midrule
    \multicolumn{7}{l}{\emph{(time constant)}} \\
    \num{1e-8} & 5 & \num{1e-2} & \num{1e-2} & 2 & \num{16.38 \pm 2.22} & \num{85.87 \pm 0.05} \\
    \num{1e-8} & 10 & \num{1e-2} & \num{1e-2} & 2 & \num{15.71 \pm 2.45} & \num{85.92 \pm 0.03} \\
    \num{1e-8} & 20 & \num{1e-2} & \num{1e-2} & 2 & \num{14.64 \pm 2.35} & \num{84.75 \pm 1.03} \\
    \num{1e-8} & 30 & \num{1e-2} & \num{1e-2} & 2 & \num{13.67 \pm 2.32} & \num{85.70 \pm 0.06} \\
    \num{1e-8} & 40 & \num{1e-2} & \num{1e-2} & 2 & \num{13.02 \pm 2.24} & \num{85.63 \pm 0.14} \\
    \num{1e-5} & 5 & \num{1e-2} & \num{1e-2} & 2 & \num{84.81 \pm 0.11} & \num{67.73 \pm 2.34} \\
    \num{1e-5} & 10 & \num{1e-2} & \num{1e-2} & 2 & \num{83.67 \pm 0.12} & \num{39.99 \pm 5.15} \\
    \num{1e-5} & 20 & \num{1e-2} & \num{1e-2} & 2 & \num{82.50 \pm 0.14} & \num{23.99 \pm 3.20} \\
    \num{1e-5} & 30 & \num{1e-2} & \num{1e-2} & 2 & \num{81.41 \pm 0.17} & \num{18.18 \pm 0.71} \\
    \num{1e-5} & 40 & \num{1e-2} & \num{1e-2} & 2 & \num{80.16 \pm 0.23} & \num{15.80 \pm 0.76} \\
    \multicolumn{7}{l}{\emph{(dyn. conv. tol.)}} \\
    \num{1e-5} & \num{30} & \num{1e-3} & \num{1e-2} & 2 & \num{81.49 \pm 0.17} & --- \\
    \num{1e-8} & \num{30} & \num{1e-3} & \num{1e-2} & 2 & --- & \num{85.54 \pm 0.32} \\
    \num{1e-5} & \num{30} & \num{1e-2} & \num{1e-2} & 2 & \num{81.46 \pm 0.21} & --- \\
    \num{1e-8} & \num{30} & \num{1e-2} & \num{1e-2} & 2 & --- & \num{85.77 \pm 0.12} \\
    \num{1e-5} & \num{30} & \num{1e-1} & \num{1e-2} & 2 & \num{54.07 \pm 1.77} & --- \\
    \num{1e-8} & \num{30} & \num{1e-1} & \num{1e-2} & 2 & --- & \num{78.57 \pm 0.65} \\
    \multicolumn{7}{l}{\emph{(learning rate sweep)}} \\
    \num{1e-9} & 30 & \num{1e-2} & \num{1e-2} & 2 & \num{13.03 \pm 2.71} & \num{75.44 \pm 0.45} \\
    \num{1e-8} & 30 & \num{1e-2} & \num{1e-2} & 2 & \num{19.52 \pm 1.40} & \num{85.73 \pm 0.08} \\
    \num{1e-7} & 30 & \num{1e-2} & \num{1e-2} & 2 & \num{60.45 \pm 0.26} & \num{52.47 \pm 6.59} \\
    \num{1e-6} & 30 & \num{1e-2} & \num{1e-2} & 2 & \num{79.54 \pm 0.14} & \num{15.94 \pm 0.32} \\
    \num{1e-5} & 30 & \num{1e-2} & \num{1e-2} & 2 & \num{83.27 \pm 0.18} & \num{17.67 \pm 1.36} \\
    \num{1e-4} & 30 & \num{1e-2} & \num{1e-2} & 2 & \num{72.71 \pm 1.16} & \num{14.58 \pm 1.22} \\
    \num{1e-3} & 30 & \num{1e-2} & \num{1e-2} & 2 & N/A & \num{11.84 \pm 1.81} \\
    \multicolumn{7}{l}{\emph{(nudge conv. tol.)}} \\
    \num{1e-5} & 30 & \num{1e-2} & \num{1e-3} & 2 & \num{41.45 \pm 16.44} & --- \\
    \num{1e-8} & 30 & \num{1e-2} & \num{1e-3} & 2 & --- & \num{85.35 \pm 0.14} \\
    \num{1e-5} & 30 & \num{1e-2} & \num{1e-2} & 2 & \num{52.05 \pm 14.96} & --- \\
    \num{1e-8} & 30 & \num{1e-2} & \num{1e-2} & 2 & --- & \num{81.25 \pm 3.16} \\
    \num{1e-5} & 30 & \num{1e-2} & \num{1e-1} & 2 & \num{79.48 \pm 2.22} & --- \\
    \num{1e-8} & 30 & \num{1e-2} & \num{1e-1} & 2 & --- & \num{84.19 \pm 0.06} \\
    \multicolumn{7}{l}{\emph{(\# layers)}} \\
    \num{1e-5} & 30 & \num{1e-2} & \num{1e-2} & 1 & \num{82.25 \pm 0.13} & --- \\
    \num{1e-8} & 30 & \num{1e-2} & \num{1e-2} & 1 & --- & \num{84.49 \pm 0.47} \\
    \num{1e-5} & 30 & \num{1e-2} & \num{1e-2} & 2 & \num{81.28 \pm 0.22} & --- \\
    \num{1e-8} & 30 & \num{1e-2} & \num{1e-2} & 2 & --- & \num{85.72 \pm 0.13} \\
    \num{1e-5} & 30 & \num{1e-2} & \num{1e-2} & 3 & \num{79.92 \pm 1.06} & --- \\
    \num{1e-8} & 30 & \num{1e-2} & \num{1e-2} & 3 & --- & \num{82.49 \pm 0.92} \\
    \bottomrule
    \end{tabular}
\end{table}

\section{Broader Impact}
\label{sec:impact}

Our work is focused on making a theoretical link between two biologically plausible learning frameworks, and therefore has limited societal implications. However, biologically plausible learning rules, in general, contribute to work enabling energy-efficient training of artificial intelligence (AI) models. On one hand, this can enable training of state-of-the-art models, such as large language models, with limited hardware resources, allowing more individuals to access this technology and guide its development. On the other hand, more accessible training for AI models could expand misuse and documented harms caused by this technology.

\end{document}